\documentclass[journal]{IEEEtran}
\usepackage{cite}
\usepackage[mathscr]{euscript}

\usepackage{graphicx}
\usepackage[cmex10]{amsmath}
\usepackage[ruled]{algorithm2e}
\usepackage{algorithmic}
\usepackage{array}
\newcolumntype{P}[1]{>{\centering\arraybackslash}p{#1}}
\newcolumntype{M}[1]{>{\centering\arraybackslash}m{#1}}
\usepackage{mdwmath}
\usepackage{mdwtab}
\usepackage{url}
\usepackage[english]{babel}
\usepackage{amsfonts}
\usepackage{mathbbol}
\usepackage{amssymb}
\usepackage{amsthm}
\usepackage{mathtools}
\usepackage{adjustbox}
\usepackage{comment}
\usepackage{multicol}
\usepackage{multirow}
\usepackage{placeins}
\usepackage{dsfont}
\usepackage{soul,color}
\usepackage{tabu}
\usepackage{multirow}
\usepackage{booktabs}
\usepackage{cleveref}
\usepackage{cite}
\usepackage{paracol}
\usepackage[many]{tcolorbox}
\usepackage{xcolor}
\usepackage{physics}

\NewTColorBox{NewBox}{ s O{!htbp}  }{%
  floatplacement={#2},
  IfBooleanTF={#1}{float*,width=\textwidth}{float},
  colframe=black!50!black,colback=white!95!black,valign=center, halign=flush center,enhanced,attach boxed title to top center={yshift=-3mm,yshifttext=-1mm},colbacktitle=black!70!white,
  title=Objectives,fonttitle=\bfseries,
  boxed title style={size=small,colframe=black!50!black}% any tcolorbox options here
  }
\usepackage{float}
\renewcommand\appendixname{Derivation of the proposed dynamical system model:}

\SetAlgoCaptionLayout{centerline}

\begin{document}

\title{A Unified Perspective of Evolutionary Game Dynamics Using Generalized Growth Transforms}

\author{Oindrila~Chatterjee,~\IEEEmembership{Student~Member,~IEEE,}
        and~Shantanu~Chakrabartty,~\IEEEmembership{Senior~Member,~IEEE}% <-this % stops a space

\thanks{Both the authors are with the Department of Electrical and Systems Engineering, Washington University in St. Louis, St. Louis, Missouri 63130, USA. All correspondences regarding this manuscript should be addressed to shantanu@wustl.edu.}
\thanks{This work was supported in part by research grants from the National Science Foundation (ECCS:1550096).}}

\maketitle
\begin{abstract}

In this paper, we show that different types of evolutionary game dynamics are, in principle, special
cases of a dynamical system model based on our previously reported framework of generalized growth transforms. The framework shows that different dynamics arise as a result of minimizing a population energy such that the population as a whole evolves to reach the most stable state.  
By introducing a population dependent time-constant in the generalized growth transform model,
the proposed framework can be used to explain a vast repertoire of evolutionary dynamics, including some novel forms of game  dynamics with non-linear payoffs.

%We extend the generalized
%growth transforms to include adaptive time-constants In particular, for strictly stable games the stable state is the Nash Equilibrium of the game, and is also globally asymptotically stable. The framework can be
%extended to non-linear domains and to a vast repertoire of evolutionary dynamics, including some novel forms of game 
%dynamics with non-linear payoffs.

%We present an extension of the growth transform dynamical system to certain forms of nonlinear domains in the course of the discussion. We also show
%how a vast repertoire of evolutionary dynamics can be arrived at by optimizing different types of network level objective functions defined over a regular simplex, or over certain types of nonlinear domains which can be mapped to a simplex. The generality of the framework also allows for the generation of novel forms of game dynamics using different forms of time constants and manifolds in the original dynamical system model. While the discretized form of the replicator dynamics under linear payoffs have been observed to be a special case of the Baum-Eagon
%inequality in earlier work, we extend the continuous time growth transform framework to encompass other types of population dynamics, and also under nonlinear payoff scenarios.

\end{abstract}

\begin{IEEEkeywords}
Dynamical systems, evolutionary game theory, growth transforms, Baum-Eagon inequality, optimization.

\end{IEEEkeywords}

\section{Introduction}
\label{intro}

\IEEEPARstart{E}{volutionary} games utilize classical game theoretic concepts to describe how a given population evolves over time as a result of interactions between the members of the population. The fitness of an individual in the population is governed by the nature of these interactions, and in accordance with the Darwinian tenets of evolution\cite{nowak2004evolutionary,hofbauer2003evolutionary}. For instance, in a strategic game each individual receives a payoff or gain according to the survival strategy it employs, as a result of which the traits or strategies with the maximum payoffs eventually dominate the population through reproduction, mutation, selection or cultural imitation.
These principles of evolutionary game dynamics have been applied to different applications ranging from genetics, social
networks, neuroeconomics to congestion control and wireless communications~\cite{hofbauer2003evolutionary,page2002unifying,basanta2008evolutionary,frey2010evolutionary,zomorrodi2017genome,han2012game,friedman1998economic,byde2003applying,luazuaroiu2017can,glimcher2004neuroeconomics,camerer2003strategizing,sugrue2005choosing}.
 
%The applications of evolutionary game dynamics are widespread,  ranging from modeling various principles of evolutionary biology like kin selection, evolutionary genetics, mating, social foraging \cite{hofbauer2003evolutionary,page2002unifying}, modeling cell proliferation and apoptosis in tumor cells \cite{basanta2008evolutionary}, behavior of microbial communities \cite{frey2010evolutionary,zomorrodi2017genome} on one hand to congestion control, routing and resource allocation, TCP throughput adaptation etc. in wireless communication networks\cite{han2012game} , auction and bidding strategies in economics \cite{friedman1998economic,byde2003applying} etc. on the other. More recently, these concepts have been applied to modeling and analyzing social networks and also in neuroeconomics, which provides a bottom-up mechanistic insight of decision-making at the cellular level, thus accounting for both physiological and behavioral factors at the root of any cognitive decision \cite{luazuaroiu2017can,glimcher2004neuroeconomics,camerer2003strategizing,sugrue2005choosing}.
\begin{figure}
\begin{center}
\includegraphics[page=2,scale=0.7,trim=2 2 2 2,clip]{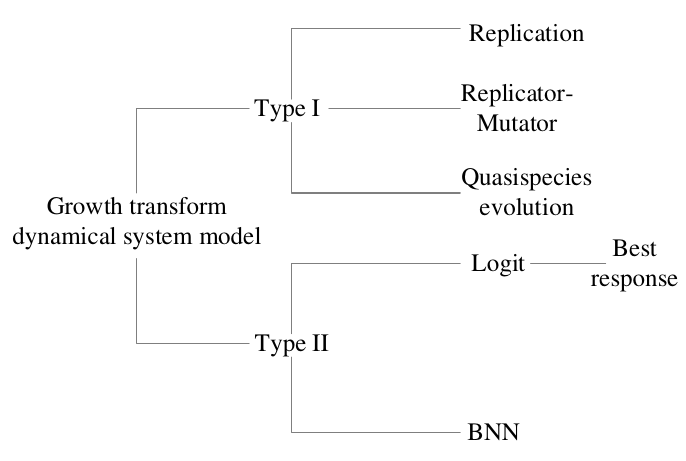}
\end{center}
%\vspace{-0.5cm}
\caption{Illustration of the scope of the paper: Most of the commonly known evolutionary stable game dynamics can be obtained from the generalized growth transform dynamical system model, by using different forms of cost functions, adaptive time constants and manifolds.}
%\vspace{-0.7cm}
%\setlength{\abovecaptionskip}{-40pt}
%\setlength{\belowcaptionskip}{-20pt}
\label{fig_main}
\end{figure}

\par In literature, numerous mathematical models have been proposed to describe the evolutionary process  \cite{nisan2007algorithmic,cohen2009evolutionary,sandholm2009evolutionary,hofbauer2003evolutionary,allen2017evolutionary}
and are briefly summarized in Figure~\ref{fig_main}. Each of these models subsumes different set of assumptions on the types of strategies and payoffs that the population employs in the course of evolution. 
%Section \ref{sec_evgames} presents some of the most common evolutionary dynamics reported till date. 
However, all of these formulations can be mathematically expressed as a dynamical system model which for stable games eventually converges to a stationary point on the probability simplex defined by the available strategies in the population. The dynamics are inherently nonlinear in the most generic case, and the existence of Nash equilibria(NE) or of Evolutionarily Stable States(ESSs) depends on the nature of the inter-species interactions\cite{li2016computing,mihai2017approximation,hofbauer2003evolutionary}. In this paper, we report that many of
these evolutionary game models are special cases of our previously reported framework of a growth transform based dynamical system model~\cite{chatterjeedyn}, as illustrated in Figure~\ref{fig_main}. The framework was derived from an energy minimization-based perspective that exploited the inherent tendency of all natural systems to evolve to their most stable state (which is also the state where the system has the minimum free energy) subject to different conservation constraints \cite{feynman2017character,friston2010free}. In this paper, we can extend the same argument to a population/network where individuals interact with each other to reach the population level most stable configuration in an energy landscape. The approach is thus similar in flavor to the concept of congestion games, and more generally potential games, which map the Nash equilibria of a game to the locally optimal points of a corresponding Lyapunov or potential function. For stable potential games involving strictly  concave potentials, a unique Nash equilibrium always exists \cite{monderer1996potential,milchtaich1996congestion,hofbauer2009stable,sandholm2009evolutionary}. 
%\par In our previous work \cite{chatterjeedyn}, we have shown how a generalized dynamical system modeled on growth transforms \cite{baum1968growth} framework can be essentially mapped to an optimization problem involving a generic Lipschitz continuous cost function over a probabilistic domain. 
Section \ref{sec_growth} gives an overview of the generalized growth transform dynamical system model. In this paper, we demonstrate how most of the commonly studied evolutionary dynamics are special cases of the growth transform based dynamical system model. Growth transforms have been applied previously for solving replicator equations with linear payoffs and in discrete-time scenarios \cite{pelillo1999replicator,bomze2013game,bomze1999maximum,nisan2007algorithmic}. In Section \ref{sec_evgames} we demonstrate how different types of known evolutionary dynamics like replicator dynamics with nonlinear payoffs, imitation games, best response dynamics, Brown-von-Neumann dynamics etc. can all be derived from the generic growth transform dynamical system model, in addition to generation of several interesting types of hitherto unexplored ones. 
%The framework is also capable of explaining adaptive evolutionary dynamics using the continuous-space formulation of the model.

%\begin{figure}
%\begin{center}
%\includegraphics[page=1,scale=0.45,trim=4 4 4 4,clip]{Images/motivation.pdf}
%\end{center}
%%\vspace{-0.5cm}
%\caption{Illustration of the scope of the paper: Each multidimensional variable/data vector is mapped into an individual growth transform limit cycle oscillator having a particular frequency, and the coupling between them leads to a constant phase difference between each pair of oscillators in steady state. In the frequency domain , this translates to a sudden shift of the superposition of all the individual waveforms from the baseline frequency during the transient state, after which the network frequency is restored to its baseline value during the steady state. }
%%\vspace{-0.7cm}
%%\setlength{\abovecaptionskip}{-40pt}
%%\setlength{\belowcaptionskip}{-20pt}
%\label{fig_motivation}

\section{Main Result}\label{sec_growth}

In \cite{chatterjeedyn}, we used the Baum-Eagon inequality\cite{baum1968growth} to show that an optimal point of a generic Lipschitz continuous cost function $H(\mathbf{p}) , \mathbf{p} \in D\subset \mathbb{R}^N$, where $D=\{\mathbf{p}\in \mathbb{R_+}^N :p_{i} \geq 0,\quad \sum\limits_{i=1}^{N} p_{i}=1\}$, corresponds to the steady state solution of the generalized growth transform dynamical system model with time-constant $\tau > 0$, given by

\begin{equation}
\tau \dot{p}_i+p_i=\dfrac{p_i\Bigg[-\dfrac{\partial H(\mathbf{p})}{\partial p_i}+\lambda \Bigg]}{\sum \limits_i p_i\Bigg[-\dfrac{\partial H(\mathbf{p})}{\partial p_i}+\lambda \Bigg]}, \label{eq_growthdyn1}
\end{equation} 
 The above equation can be rewritten in a compact form as follows
\begin{equation}
\tau \dot{p}_i = p_i\Bigg[ \dfrac{f_i(\mathbf{p})}{\bar{f}}-1\Bigg],
\label{eq_growthdyn2}
\end{equation}  
where $f_i(\mathbf{p})=\Bigg[-\dfrac{\partial H(\mathbf{p})}{\partial p_i}+\lambda \Bigg] \quad \forall i$ and $\bar{f}= \sum_i p_if_i(\mathbf{p})$. The constant $\lambda \in \mathbb{R}_+$ is chosen to ensure that 
$f_i(\mathbf{p}) > 0, \forall i$. 

%The dynamical system formulation was arrived at by using the homotopically increasing property of the Baum-Eagon inequality\cite{baum1968growth}. These growth transforms originally employed for maximizing an arbitrary polynomial $H(\mathbf{p})$ with real coefficients over a probabilistic domain, they were later extended for maximization of rational \cite{gopalakrishnan1989generalization} and analytic functions over more generic nonlinear sub-manifolds of the probabilistic domain \cite{kanevsky2004extended}, and even for accommodating box constraints on the optimization variables \cite{gangopadhyay2018extended}. 

\par However, the convergence of Equation~(\ref{eq_growthdyn1}) to the steady-state solution also holds if the time-constant $\tau$ is also varied with time. This is because the normalization constraint is the only condition on the population state at any instant of time so as to ensure that the manifold $D$ is an invariant manifold, and hence other types of time constants can also be used, provided it is identical for all the strategies. If different strategies evolve according to different time constants, however, convergence to a locally optimal solution for Lipschitz continuous cost functions is not guaranteed. In this paper we will vary $\tau$ as a function of $\mathbf{p}$ such that $\tau =\dfrac{1}{\bar{g}(\mathbf{p})}$, where $\bar{g}(\mathbf{p}) > 0$ could be an arbitrary time-varying function. Substituting for $\tau$ in Equation~(\ref{eq_growthdyn2}) we arrive at the key dynamical system model equation:

%Though in \cite{chatterjeedyn}, the time constant $\tau$ was treated as a constant variable, it can be easily observed that the model introduced in the previous section also allows for an adaptive time constant $\tau(t)$, or in other words, an adaptive step size, since this would still satisfy the normalization constraint over a regular $n-$simplex at any instant of time. For example, we can have a time constant of the form $\tau =\dfrac{1}{\bar{g}(\mathbf{p})}$, where $\bar{g}(\mathbf{p})$ can be any arbitrary time varying function, which has the same form for all the variables of interest. This would lead to the following dynamical system model:
 
\begin{equation}
\dot{p}_i =p_i \Bigg[ \dfrac{f_i(\mathbf{p})}{\bar{f}}\bar{g}(\mathbf{p})-\bar{g}(\mathbf{p})\Bigg].
\label{eq_growthdyn3}
\end{equation}

We will now show that Equation~(\ref{eq_growthdyn3}) can be used to explain different evolutionary game dynamics reported in literature and also for proposing some new forms of dynamics.

%We can thus arrive at different types of interesting dynamics by choosing different forms of $g(\cdot)$ as well, in addition to choosing different forms of cost functions or domains.

\begin{figure}
\begin{center}
\includegraphics[page=2,scale=0.38,trim=4 4 4 4,clip]{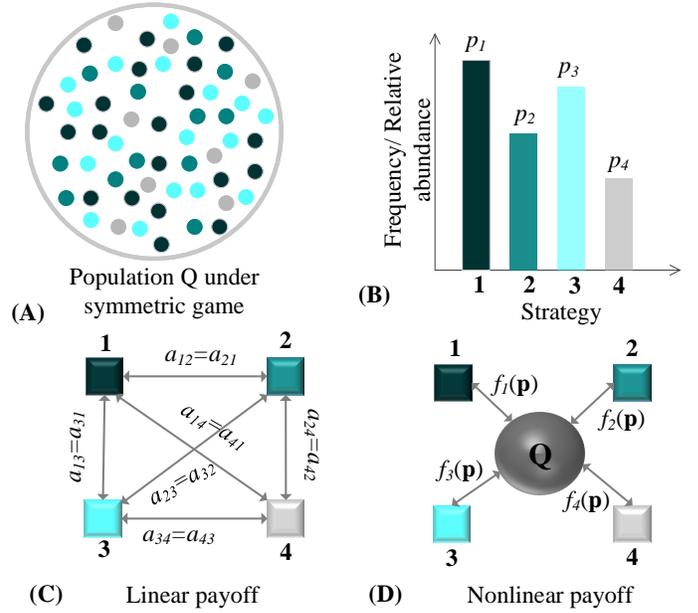}
\end{center}
%\vspace{-0.5cm}
\caption{Illustration of a symmetric game setup with four pure strategies: (a)Population at any instant of time (colors denote the strategy each individual adopts at any time $t$); (b)Histogram showing the relative abundance of the strategies in the population, $p_1-p_4$; (c)Linear payoff case, where the payoff of an individual with strategy $i$ playing against another with startegy $j$ is given by $a_{ij}$, with the fitness of $i$ being $\sum\limits_j a_{ij}p_j$; (d)Nonlinear payoff case, where each strategy $i$ has a fitness $f_i(\mathbf{p})$ against the population $Q$.}
%\vspace{-0.7cm}
%\setlength{\abovecaptionskip}{-40pt}
%\setlength{\belowcaptionskip}{-20pt}
\label{fig_gamesetup}
\end{figure}

\section{Evolutionary Game Dynamics}
\label{sec_evgames}

%\par In this section, we show how by considering an adaptive time constant which is inversely proportional to the instantaneous mean payoff of the population, and using different forms of the cost function $H$ and domain of definition $D$, we show how different types of evolutionary game dynamics can be arrived at by using the growth transform dynamical system model. 

We first introduce some mathematical notations and terminology pertaining to evolutionary dynamics arising in the context of  deterministic games \cite{hofbauer2003evolutionary,page2002unifying,nisan2007algorithmic}.
In deterministic games, the Nash equilibrium (if it exists), remains unchanged even when perturbed by a small percentage of mutants or external agents playing a different strategy. Also, we will only consider stable games that converge to a global and stable Nash equilibrium\cite{sandholm2009evolutionary,hofbauer2009stable}.  \par Consider an infinitely large population with a finite set $\mathbf{S}=\{1, \ldots,N\}$ of $N$ pure strategies available to each individual of the population. The $i^{th}$ strategy is then quantified using a relative abundance measure $p_i\in \mathbb{R}_+$ and a fitness function (or expected payoff) $f_i(\mathbf{p})$ ($f_i : \mathbb{R}_+^N \mapsto \mathbb{R}_+$), where $\mathbf{p}\in D$ is the set with elements $p_i$. It should be noted here that though the fitness is a function of the relative abundances of various strategies adopted by the population in the most generic case, constant fitness scenarios are also possible in certain types of evolutionary games. Also, for most of the dynamics discussed in this paper, we will consider symmetric games, where all individuals have the same strategy set $\mathbf{S}$ during the course of evolution. Figure \ref{fig_gamesetup}(a) shows an example of a symmetric game comprising of four pure strategies, where each color denotes the strategy that each individual in the population $Q$ adopts at any instant of time. Figure \ref{fig_gamesetup}(b) shows the relative abundance $p_i$ for each strategy in the population and Figure \ref{fig_gamesetup}(c) shows an example of a linear payoff function. In this case, each individual which plays a strategy $i$ against any random member of the population playing a strategy $j$ receives a constant payoff $a_{ij}$. The average payoff or fitness of the $i-$th strategy is then given by $\sum \limits_j a_{ij}p_j$ . Figure \ref{fig_gamesetup}(d) illustrates a more general payoff scenario, where the average payoff corresponding to the $i-th$ strategy could be a non-linear function $f_i(\mathbf{p})$. 

Next, we present some of the most common types of evolutionary game dynamics existing in literature, and their relation to the growth transform dynamical system model proposed in Equation (\ref{eq_growthdyn3}) by choosing the function $\bar{g}(\mathbf{p})$ to be directly proportional to the instantaneous mean payoff of the population, and using different forms of the cost function $H$.

\subsection{Replicator dynamics }
 Replicator dynamics \cite{schuster1983replicator} model the case where evolution occurs by pure natural selection, and the fitness of the individual is governed by the frequencies of other strategies in the population.  The population thus evolves in a way as to promote strategies having payoffs higher than the mean payoff at any instance of time. The generic form of replicator dynamics with a nonlinear fitness function is given by:
\begin{equation}
\dot{p}_i = p_i[f_i(\mathbf{p})-\bar{f}], \quad \bar{f}=\sum \limits_i p_if_i(\mathbf{p}) \label{eq_rep1}
\end{equation}
%Linear fitness functions involving a constant payoff matrix $\mathbf{A}$ in a symmetric game, where $a_{ij}$ denotes the fitness of strategy $i$ against any other strategy $j$, is given by:  
%\begin{equation}
%\dot{p}_i = p_i[(\mathbf{Ap})_i-\mathbf{p}^T\mathbf{Ap}]\label{eq_rep2}
%\end{equation}
%with $(\mathbf{Ap}_i)$ being the fitness or average payoff of strategy $S_i$, with its equivalent discretized version being
%\begin{equation}
%p_i(t+1)=p_i(t)\dfrac{\mathbf{Ap}(t)_i+\lambda}{\sum \limits_i p_i(t)(\mathbf{Ap}(t)_i+\lambda)},\label{eq_rep3}
%\end{equation}

%\begin{equation}
%p_i(t+1)=p_i(t)\dfrac{\exp(\mathbf{Ap}(t)_i)}{\sum \limits_j p_j(t)\exp(\mathbf{Ap}(t)_j)}
%\end{equation} 
%where $\lambda$ is a constant baseline payoff. 

\par The replicator dynamics can be derived from the growth transform dynamical system model by considering $\bar{g}(\mathbf{p})=\sum \limits_i p_i(f_i(\mathbf{p})+\lambda)$ and a network level cost function of the form
\begin{equation}
H(\mathbf{p})=-\sum \limits_i \int_c^{p_i}f_i(\mathbf{z})dz_i-\lambda \sum \limits_i p_i,
\end{equation}
where $f_i(\mathbf{p})$ is a smooth monotonic function of $p_i$, and $\lambda>0$ is chosen so as to ensure $f_i(\mathbf{p})>0$, and can be thought of as a constant background payoff. Note that $f_i(\mathbf{z}) = f_i(p_1,\ldots,z_i,\ldots, p_n) \quad \forall i$, where $z_i$ is a
dummy variable corresponding to the $i-$th strategy.  
%The time constant thus increases with time, leading to faster evolution towards the beginning of the process to very slow dynamics near convergence, which is also true for most natural processes.

\par Of particular interest are the replicator dynamics under a linear payoff scenario with a constant payoff matrix $\mathbf{A}$ in a symmetric game ($a_{ij}$ being the fitness of strategy $i$ against any other strategy $j$) and $(\mathbf{Ap}_i)$ being the fitness or average payoff of strategy $S_i$. This can be derived from the growth transform dynamical system model by using $\bar{g}(\mathbf{p}) = \sum_i p_i((\mathbf{Ap})_i+\lambda)$ and a cost function of the form $H(\mathbf{p}) = -\mathbf{p}^T\mathbf{Ap}-\lambda \sum \limits_i p_i$  over
the probabilistic domain $D$. The corresponding dynamics are given by: 
\begin{equation}
\dot{p}_i = p_i[(\mathbf{Ap})_i-\mathbf{p}^T\mathbf{Ap}]\label{eq_rep2}
\end{equation}
 In \cite{pelillo1999replicator,bomze2000approximating}, the discrete-time version of Equation (\ref{eq_rep2}) given by
\begin{equation}
p_i(t+1)=p_i(t)\dfrac{\mathbf{Ap}(t)_i+\lambda}{\sum \limits_i p_i(t)(\mathbf{Ap}(t)_i+\lambda)},\label{eq_rep3}
\end{equation}
where $\lambda$ is a constant baseline payoff, was shown to be a special case of the Baum-Eagon inequality, and was employed for solving maximum clique and graph isomorphism problems which are commonly encountered in computer vision \cite{bomze1999maximum,pelillo1999replicator,bomze2013game,mehta2017mutation}.

\par The Lotka-Volterra equation can also be derived from the linear payoff form of the replicator equation by means of a simple linear transformation.

\subsection{Quasispecies evolution } 
These dynamics correspond to a constant fitness landscape \cite{eigen1989molecular}, where replications from one generation to the next are error-prone with a high enough mutation rate, i.e., the $j-$th strategy can mutate to the $i-$th strategy with a probability $m_{ji}$. Forward and backward mutations are equally probable in this model, i.e., the mutation matrix $\mathbf{M} = \{m_{ij}\}$ is a doubly stochastic symmetric matrix, with $m_{ij} = m_{ji}$, and $\sum \limits_i m_{ij}=\sum \limits_j m_{ij}=1$. The dynamical system
corresponding to this model of evolution is as follows  :
\begin{equation}
\dot{p}_i=\sum \limits_j p_jf_jm_{ji}-p_i\sum\limits_i p_if_i, \quad \sum \limits_i m_{ji}=1 \enskip \forall i \label{eq_quasi}
\end{equation}
\par Noting that the mutation matrix
$\mathbf{M}$ is doubly stochastic and that each strategy has a constant
fitness, the quasispecies dynamics can be achieved by using $\bar{g}(\mathbf{p})=\sum \limits_i p_i[f_i+\lambda]$ and a cost function of the form
\begin{equation}
H(\mathbf{p})=-\sum \limits_i \log(p_i)\sum \limits_j p_jf_jm_{ji}-\lambda\sum \limits_i p_i,
\end{equation}

since $\sum \limits_i m_{ji}=1$.

\subsection{Replicator-mutator dynamics }
When both replication and mutation contribute to the evolutionary process, we arrive at the replicator-mutator equation \cite{hadeler1981stable}:
\begin{equation}
\dot{p}_i=\sum \limits_j p_jf_j(\mathbf{p})m_{ji}-p_i\sum \limits_i p_i f_i(\mathbf{p}), \quad \sum \limits_i m_{ji}=1 \enskip \forall i \label{eq_repmut}
\end{equation} 

\par In a fashion similar to the previous two types of evolutionary dynamics, the
mathematical equation governing the replicator-mutator
model can be arrived at by considering and time constant $\bar{g}(\mathbf{p})=\sum \limits_i p_i[f_i(\mathbf{p})+\lambda]$ and a cost function of the form
\begin{equation}
H(\mathbf{p})=-\sum \limits_i \int_c^{p_i} \sum \limits_j p_jf_j(\mathbf{z})m_{ij}dz_i-\lambda\sum \limits_i p_i.
\end{equation}

The above three examples belong to a more general class of evolutionary dynamics called \textit{imitation games}, which are of the form $\dot{p}_i=p_i\zeta_i(\mathbf{p})$, with $\sum_i p_i\zeta_i(\mathbf{p})=0 \quad \forall \mathbf{p} \in D$.

\subsection{Logit dynamics and best response dynamics } 
The logit dynamics \cite{alos2010logit} represent a class of evolutionary games where the individuals have a partial knowledge about the global state of the system, with a parameter $\eta>0$ representing the noise level of the system. At each step, an individual selects a strategy so as to select the current best response, depending on the noisy population-level knowledge that it possesses. The dynamics of this form are given by:
\begin{equation}
\dot{p}_i=\dfrac{\exp(f_i(\mathbf{p})/{\eta})}{\sum \limits_i \exp(f_i(\mathbf{p})/{\eta})}-p_i \label{eq_logit}
\end{equation}
For noiseless systems, i.e., when $\eta \to 0$, this converges to the best response dynamics \cite{matsui1992best}, where each individual chooses the best available strategy with the highest probability. A discrete time version of the best response dynamics is the fictitious play \cite{monderer1996fictitious}.

\par For arriving at the Logit dynamics from the growth transform model, we choose $\bar{g}(\mathbf{p})=\sum \limits_i \exp(f_i(\mathbf{p})/\eta)$ a cost function of the form:
\begin{equation}
 H(\mathbf{p})=-\sum \limits_i \int_c^{p_i}\dfrac{1}{z_i}\exp{f_i(\mathbf{z}/\eta)} dz_i, \quad p_i>0 \enskip \forall i, 
\end{equation}
where $f_i(\mathbf{z})$ has the same definition as in the replicator dynamics case.

\subsection{Brown-von Neumann-Nash (BNN) dynamics }
When each strategy gets updated only if its payoff is greater that the average payoff of the population with a certain margin $\epsilon$, we obtain the following equation which corresponds to the Brown-von Neumann-Nash dynamics \cite{hofbauer2009brown}:
\begin{equation}
\dot{p}_i=k_i(\mathbf{p})-p_i\sum \limits_i k_i(\mathbf{p}) \label{eq_bnn}
\end{equation}
where $k_i(\mathbf{p})=\max(0,((\mathbf{Ap})_i-\mathbf{p}^T\mathbf{Ap}+\epsilon))$.

\par The BNN dynamics, similarly, can be achieved by
considering $\bar{g}(\mathbf{p})=\sum \limits_i k_i(\exp(\mathbf{x}))$ and a cost function of the form:
\begin{equation}
H(\mathbf{p})= -\sum\limits_i \int_c^{p_i} \dfrac{1}{z_i} k_i(\mathbf{z})dz_i
\end{equation}
%\begin{equation}
%H(\mathbf{x})=-\sum\limits_i \int_c^{x_i} k_i(\exp(\mathbf{z}))dz_i
%\end{equation}
where $k_i(\mathbf{p})=\max(0,((\mathbf{Ap})_i-\mathbf{p}^T\mathbf{Ap}+\epsilon))$ as before, and $p_i>0 \enskip \forall i$.

\subsection{Unconventional and novel dynamics } \label{subsec_novel}

\par Novel types of evolutionary dynamics can be arrived at by using other forms of the function $\bar{g}(\mathbf{p})$, and also by using other variants of the cost function. A particularly interesting scenario uses a cost function $H(\mathbf{p})=-\sum \limits_i \int_c^{p_i}h(z_i)f_i(\mathbf{z})dz_i$, which leads to a dynamical system of the form:
\begin{equation}
\dot{p}_i =p_i \Bigg[ \dfrac{f_i^\prime(\mathbf{p})}{\bar{f^\prime}}\bar{g}(\mathbf{p})-\bar{g}(\mathbf{p})\Bigg],
\label{eq_growthdynnov}
\end{equation}
where $f_i^\prime(\mathbf{p})=h(p_i)f_i(\mathbf{p})=h(p_i)\Bigg[-\dfrac{\partial H}{\partial p_i}+\lambda\Bigg] \quad \forall i$.

We can choose a form of $h(p_i) \quad \forall i$ such that only a certain sub-population is selected to determine the mean fitness of the entire population. Potential candidates for the form of the functions $h(\cdot)$ include derivatives of smooth saturation functions like the hyperbolic tangent function, the logistic function, the sigmoidal function, the softmax function etc. For example, using $h(p_i)=\sech^2(p_i)$ (which corresponds to the derivative of the hyperbolic tangent function) in Equation (\ref{eq_growthdyn3})
 would lead to $f_i^\prime(\mathbf{p})=\sech^2(p_i)f_i(\mathbf{p})$,
%\begin{equation}
%f_i^\prime(\mathbf{p})=[1-(\tanh^2(p_i))]f_i(\mathbf{p}),
%\end{equation}
which updates only certain members of the population instead of the entire population, as in the case of the standard evolutionary games. Though such types of optimization problems do not necessarily have a globally optimal solution due to the non-convexity of the domain and/or the objective function under consideration, convergence to a locally optimal solution is always guaranteed in such cases. Intuitively, this implies that even when each strategy in the population gets updated with limited information about the overall population behavior, the population will converge to a locally stable solution eventually. The concept is thus similar in essence to the lattice based population dynamics \cite{lieberman2005evolutionary,nowak2010evolutionary} approach, where each individual updates its strategy based only on the strategies of a certain number of players in its local neighborhood, which can be described by a graph or by a lattice structure. Such neighborhood based models which allow local interactions and dispersal and operate at relatively smaller spatial scales were shown to promote coexistence of strategies leading to biodiversity \cite{kerr2002local}, since dominated strategies in such scenarios can form isolated clusters or can survive by moving over time to a spatially different location in the lattice structure \cite{hofbauer2003evolutionary}.

%\section{On the convergence to a Nash equilibrium and evolutionarily stable strategies}\label{subsec_convergence}

%\section{Experimental Results}
%\label{sec_expres}

\section{Discussions and Conclusions}
\label{sec_conclusions}

In this paper, we used an energy minimization framework based on the generalized growth transform dynamical system model to explain different types of network dynamics for stable evolutionary games having a unique strictly stable Nash equilibrium. We also showed how unexplored evolutionary processes can also be mapped to a growth transform dynamical system with an adaptive time constant, using different variants of the underlying cost function and domain of definition. In this regard the optimal point of the objective function can be related to the Nash equilibrium and the evoluationaly stable state (ESS) of the evolutionary game.
 
\par A Nash equilibrium of a game corresponds to the set of strategies adopted by a population such that any deviation from it in the subsequent stages would not result in an increased payoff against the current strategy. An ESS, on the other hand, is a population state which is inherently stable in the sense that it cannot be invaded by a small group of mutants playing the same strategy. While an ESS is always a Nash equilibrium, the converse is not necessarily true, and only holds for strict Nash equilibria \cite{hofbauer2003evolutionary,li2016computing}. In the context of the growth transform based formulation, we can conclude the following correspondence between the cost function $H$ and the nature of the equilibria:
\begin{itemize}
\item A generic Lipschitz continuous cost function over a convex domain might have multiple local minima, which correspond to the Nash equilibria of the game, and do not necessarily imply evolutionary stability.
\item A general convex cost function over a convex domain has a set of Nash equilibria which are always stable for the types of dynamics discussed in the paper, and globally stable for BR and BNN dynamics in particular.
\item A strictly convex objective function over a convex domain has a unique strict Nash equilibrium which is also the ESS, and is globally stable for all types of evolutionary dynamics discussed in this paper.
\item A strictly concave cost function over a convex domain leads to a locally or globally repelling Nash equilibrium.
\end{itemize}

Future directions will involve extension of the proposed dynamical system framework for incorporate continuous strategy spaces leading to adaptive dynamics, and scenarios involving multiple types of equilibria. 

%\nocite{*}
\bibliographystyle{ieeetr}
\bibliography{references}
\end{document}